\documentclass[runningheads]{llncs}

\usepackage{graphicx}
\usepackage{amsmath,amssymb} 
\usepackage{epsfig}
\usepackage{bm}
\usepackage[american]{babel}
\usepackage{booktabs}       
\usepackage{amsfonts}       
\usepackage{microtype}      
\usepackage{multirow}
\usepackage{color}
\usepackage{algorithm,algorithmic}
\usepackage{subfig}
\usepackage{wrapfig}

\graphicspath{{figure/}}
\begin{document}

\title{Coarse-to-fine: A RNN-based hierarchical attention model for vehicle re-identification\thanks{The first two authors contributed equally to this work. This research was supported by NSFC (National Natural Science Foundation of China) under 61772256. Liu's participation was in part supported by ARC DECRA Fellowship (DE170101259).}} 
\titlerunning{RNN-HA for Vehicle Re-ID} 


\author{Xiu-Shen Wei\inst{1}\orcidID{0000-0002-8200-1845} \and
Chen-Lin Zhang\inst{2}\orcidID{0000-0002-3168-1852} \and
Lingqiao Liu\inst{3}\orcidID{0000-0003-3584-795X} \and
Chunhua Shen\inst{3}\orcidID{0000-0002-8648-8718} \and
Jianxin Wu\inst{2}\orcidID{0000-0002-2085-7568}
}
%

\authorrunning{X.-S. Wei \emph{et al.}} 


\institute{Megvii Research Nanjing, Megvii Technology Ltd. (Face++), China \and
National Key Laboratory for Novel Software Technology, Nanjing University, China \and
School of Computer Science, The University of Adelaide, Australia
\email{weixiushen@megvii.com, \{zhangcl, wujx\}@lamda.nju.edu.cn, \\ \{lingqiao.liu, chunhua.shen\}@adelaide.edu.au}\\
}

\maketitle

\begin{abstract}
Vehicle re-identification is an important problem and becomes desirable with the rapid expansion of applications in video surveillance and intelligent transportation. By recalling the identification process of human vision, we are aware that there exists a native hierarchical dependency when humans identify different vehicles. Specifically, humans always firstly determine one vehicle's coarse-grained category, \emph{i.e.}, the car model/type. Then, under the branch of the predicted car model/type, they are going to identify specific vehicles by relying on subtle visual cues, \emph{e.g.}, customized paintings and windshield stickers, at the fine-grained level. Inspired by the coarse-to-fine hierarchical process, we propose an end-to-end RNN-based Hierarchical Attention (RNN-HA) classification model for vehicle re-identification. RNN-HA consists of three mutually coupled modules: the first module generates image representations for vehicle images, the second hierarchical module models the aforementioned hierarchical dependent relationship, and the last attention module focuses on capturing the subtle visual information distinguishing specific vehicles from each other. By conducting comprehensive experiments on two vehicle re-identification benchmark datasets VeRi and VehicleID, we demonstrate that the proposed model achieves superior performance over state-of-the-art methods.
\end{abstract}


\section{Introduction}

Vehicle re-identification is an important yet frontier problem, which aims at determining whether two images are taken from the same specific vehicle. It has diverse applications in video surveillance~\cite{iee2005}, intelligent transportation~\cite{zhangITS2011} and urban computing~\cite{acmist2014}. Moreover, vehicle re-identification has recently drawn increasing attentions in the computer vision community~\cite{fact16ICME,provid16ECCV,cuhkcar17ICCV}.

\begin{figure}[t]
\centering
	{\includegraphics[width=0.65\columnwidth]{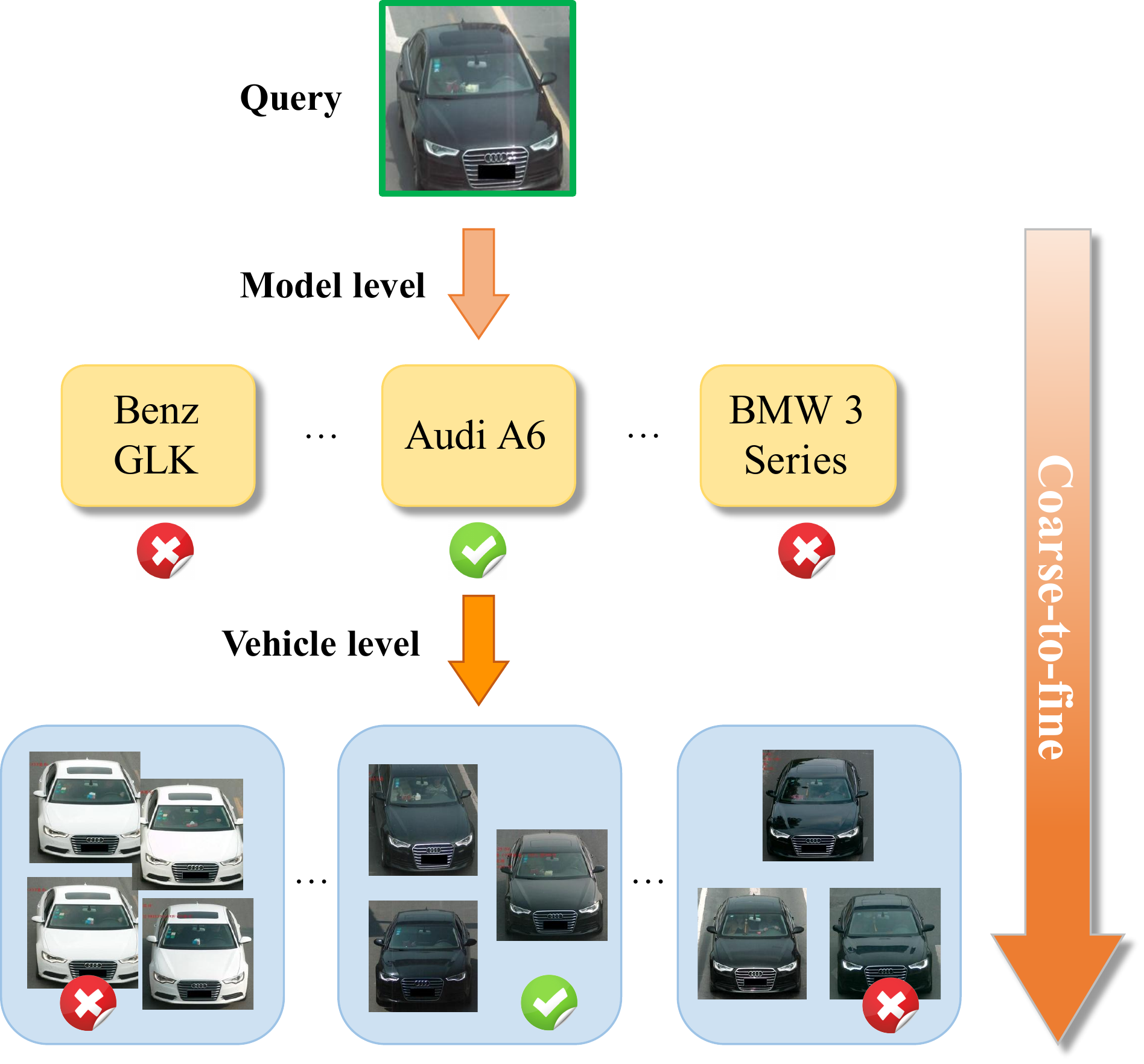}}
\caption{Illustration of coarse-to-fine hierarchical information as a latent but crucial cue for vehicle re-identification. (Best viewed in color and zoomed in.)}
\label{fig:idea}
\end{figure}

Compared with the classic person re-identification problem, vehicle re-identification could be more challenging as different specific vehicles can only be distinguished by slight and subtle differences, such as some customized paintings, windshield stickers, favorite decorations, etc. Nevertheless, there still conceals some latent but crucial information for handling this problem. As shown in Fig.~\ref{fig:idea}, when humans identify different vehicles, they always follow a \emph{coarse-to-fine} identification process. Specifically, we tend to firstly determine this specific vehicle belongs to which car model/type. The first step can eliminate many distractors, \emph{i.e.}, vehicles with similar subtle visual appearances but belonging to the other different car models/types. In the following, within the candidate vehicle set of the same car model/type, humans will carefully distinguish different vehicles from each other by using these subtle visual cues. Apparently, there is a hierarchical dependency in this coarse-to-fine process, which is yet neglected by previous studies~\cite{fact16ICME,provid16ECCV,cuhkcar17ICCV,vehicleID16CVPR}.

Motivated by such human's identification process, we propose a unified \underline{RNN}-based \underline{H}ierarchical \underline{A}ttention (RNN-HA) classification model for vehicle re-identification. Specifically, as shown in Fig.~\ref{fig:pipeline}, our RNN-HA consists of three main modules: \textit{1)} the representation learning module, \textit{2)} the RNN-based hierarchical module and \textit{3)} the attention module. The first module encodes the discriminative information of vehicle images into the corresponding deep convolutional descriptors. Then, the second RNN-based hierarchical module will mimic the coarse-to-fine identification process. Concretely, at the coarse-grained level classification (for car models), we first aggregate these deep descriptors by global average pooling (named as ``image embedding vector $\bm{x}_1$'' in Fig.~\ref{fig:pipeline}), which is expected to retain global information contributing to the coarse-grained level. We then feed the image embedding vector $\bm{x}_1$ as the input at time step 1 into the RNN-based hierarchical module. The output at time step 1 is used for the coarse-grained level classification. More importantly, the same output is also treated as the source for generating an attention guidance signal which is crucial for the subsequent fine-grained level classification. At the fine-grained level classification (for specific vehicles), to capture subtle appearance cues, we leverage an attention module where the aforementioned attention guidance signal will evaluate which deep descriptors should be attended or overlooked. Based on the attention module, the original deep descriptors are transformed into the attended descriptors which reflect those image regions containing subtle discriminative information. After that, the attended descriptors are aggregated by global average pooling into the attention embedding vector $\bm{x}_2$ (cf. Fig.~\ref{fig:pipeline}). Then, the attention embedding vector $\bm{x}_2$ is fed into RNN at time step 2 for the fine-grained level classification. In the evaluation, we employ the trained RNN-HA model as a feature extractor on test vehicle images (whose vehicle categories are disjoint with the training categories) to extract the outputs at time step 2 as the feature representations. For re-identification, these representations are first $\ell_2$-normalized, and then the cosine acts as the similarity function for computing relative distances among different vehicles.

In experiments, we perform the proposed RNN-HA model on two vehicle re-identification benchmark datasets, \emph{i.e.}, \emph{VeRi}~\cite{fact16ICME} and \emph{VehicleID}~\cite{vehicleID16CVPR}. Empirical results show that our RNN-HA model significantly outperforms state-of-the-art methods on both datasets. Furthermore, we also conduct ablation studies for separately investigating the effectiveness of the proposed RNN-based hierarchical and attention modules in RNN-HA.

In summary, our major contributions are three-fold:
\begin{itemize}
\item For vehicle re-identification, we propose a novel end-to-end trainable RNN-HA model consisting of three mutually coupled modules, especially two crucial modules (\emph{i.e.}, the RNN-based hierarchical module and the attention module) which are tailored for this problem.
\item Specifically, by leveraging powerful RNNs, the RNN-based hierarchical module models the coarse-to-fine category hierarchical dependency (\emph{i.e.}, from car model to specific vehicle) beneath vehicle re-identification. Furthermore, the attention module is proposed for effectively capturing subtle visual appearance cues, which is crucial for distinguishing different specific vehicles (\emph{e.g.}, two cars of the same car model).
\item We conduct comprehensive experiments on two challenging vehicle re-identification datasets, and our proposed model achieves superior performance over competing previous studies on both datasets. Moreover, by comparing with our baseline methods, we validate the effectiveness of two proposed key modules.
\end{itemize}


\section{Related work}\label{sec:related}

We briefly review two lines of related work: vehicle re-identification and developments of deep neural networks.

\subsection{Vehicle re-identification}

Vehicle re-identification is an important application in video surveillance, intelligent transportation and public security~\cite{fact16ICME,provid16ECCV,vehicleID16CVPR,carTMM12,compcar15CVPR}. It is a frontier research area in recent years with limited related studies in the literature.

Feris \emph{et al.}~\cite{carTMM12} proposed an attribute-based approach for vehicle search in surveillance scenes. By classifying with different attributes, \emph{e.g.}, colors and sizes, retrieval was performed on searching vehicles with such similar attributes in the database. In 2015, Yang \emph{et al.}~\cite{compcar15CVPR} proposed the large-scale car dataset \emph{CompCar} for multiple car-related tasks, such as car model classification, car model verification and attribute prediction. However, the category granularity of the studying tasks in~\cite{sochor16CVPR,compcar15CVPR} just stayed in the car model level, which was not fine enough for the specific vehicle level. Very recently, Liu \emph{et al.}~\cite{fact16ICME} and Liu \emph{et al.}~\cite{vehicleID16CVPR} proposed their vehicle re-identification dataset \emph{VeRi} and \emph{VehicleID}, respectively. After that, vehicle re-identification started to gain attentions in the computer vision community.

Specifically, the \emph{VeRi} dataset~\cite{fact16ICME} contains over $50,000$ images of $776$ vehicles captured by twenty cameras in a road network. On this dataset, \cite{fact16ICME} proposed an appearance-based method named FACT combining both low-level features (\emph{e.g.}, color and texture) and high-level semantic information extracted by deep neural networks. Later, \cite{provid16ECCV} and \cite{cuhkcar17ICCV} tried to deal with vehicle re-identification by using not only appearance information but also complex and hard-acquired spatio-temporal information. On the other hand, the \emph{VehicleID} dataset~\cite{vehicleID16CVPR} is a large-scale vehicle dataset containing about $26,000$ different vehicles with $222,000$ images. Liu \emph{et al.}~\cite{vehicleID16CVPR} proposed a mixed structured deep network with their coupled cluster loss to learn the relative distances of different vehicles on that dataset, which only depended on appearance cues.

In this paper, we attempt to use our RNN-HA model to deal with vehicle re-identification by depending on purely visual appearance cues. The reason is that appearance information is directly beneath vehicle images, and the images are easy to collect. Compared with that, complex and expensive spatio-temporal information is much harder to gather. For example, the two existing \emph{large-scale} car/vehicle datasets, \emph{e.g.}, \emph{CompCar}~\cite{compcar15CVPR} and \emph{VehicleID}~\cite{vehicleID16CVPR}, only provide the appearance-based labels such as car types, models and identified annotations. Whilst, they do not provide any spatio-temporal information. Moreover, our experimental results prove that it is practical to use only appearance information for accurate vehicle re-identification, cf.~Table~\ref{table:pkuresult}.

\subsection{Deep neural networks}

Deep convolutional neural networks (DCNNs) try to model the high-level abstractions of the visual data by using architectures composed of multiple non-linear transformations. Recent progresses in diverse computer vision applications, \emph{e.g.}, large-scale image classification~\cite{CNN12}, object detection~\cite{ssd16ECCV,chen2016accv} and semantic segmentation~\cite{Jonathan15CVPR}, are made based on the developments of the powerful DCNNs.

On the other hand, the family of Recurrent Neural Networks (RNNs) including Gated Recurrent Neural Networks (GRUs)~\cite{gru14EMNLP} and Long-Short Term Memory Networks (LSTMs)~\cite{rnn} have recently achieved great success in several tasks including image captioning~\cite{johnson16CVPR,wuqi16CVPR,imcap15ICML,tan16accv}, visual question answering~\cite{gaovqa15NIPS,wuqivqa16CVPR}, machine translation~\cite{MTL14NIPS}, etc. These works prove that RNNs are able to model the temporal dependency in sequential data and learn effective temporal feature representations, which inspires us to rely on RNNs for modeling the hierarchical coarse-to-fine characteristic (\emph{i.e.}, from car model to specific vehicle) beneath the vehicle re-identification problem.

\begin{figure*}[p]
\centering
	{\includegraphics[width=0.85\textwidth]{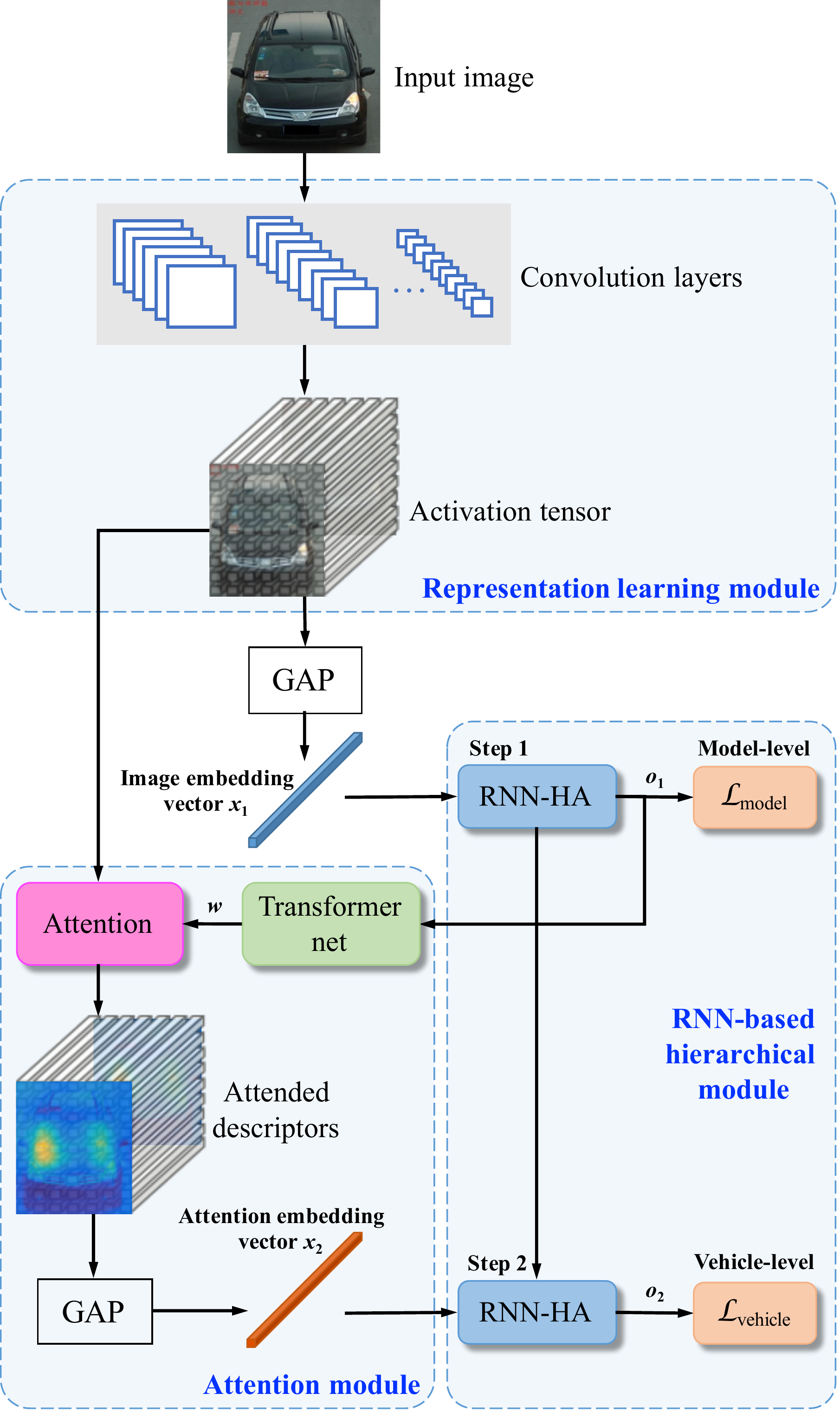}}
\caption{Framework of the proposed RNN-HA model. Our model consists of three mutually coupled modules, \emph{i.e.}, representation learning module, RNN-based hierarchical module and attention module. (Best viewed in color.)}
\label{fig:pipeline}
\end{figure*}

\section{Model}\label{sec:model}

We propose a novel RNN-based hierarchical attention (RNN-HA) classification model to solve the vehicle re-identification problem. In this section, we present our RNN-HA model by introducing its key constituent modules, \emph{i.e.}, the representation learning module, the RNN-based hierarchical module and the attention module, respectively. The framework of our model is illustrated in Fig.~\ref{fig:pipeline}.

\subsection{Representation learning module}

As shown in Fig.~\ref{fig:pipeline}, the first module of RNN-HA is to learn a holistic image representation from an input vehicle image via traditional convolutional neural networks. The obtained deep image representations will be processed by the following modules. This representation learning module consists of multiple traditional convolutional layers (equipped with ReLU and pooling). However, different from previous vehicle re-identification methods (\emph{e.g.},~\cite{vehicleID16CVPR,cuhkcar17ICCV}), we discard fully connected layers, and utilize convolutional activations for subsequent processing.

Concretely, the activations of a convolution layer can be formulated as an order-3 tensor $T$ with $h\times w\times d$ elements, which includes a set of 2-D feature maps. These feature maps are embedded with rich spatial information, and are also known to obtain mid- and high-level information, \emph{e.g.}, object parts~\cite{cross16TPAMI,singh2014accv}. From another point of view, these activations can alternatively be viewed as an array of $d$-dimensional deep descriptors extracted at $h\times w$ locations. Apparently, these deep convolutional descriptors have more local/subtle visual information and more spatial information than those of the fully connected layer features.

After obtaining the activation tensor, at the coarse-grained classification level (\emph{i.e.}, model-level), we directly apply global average pooling (GAP) upon these deep descriptors and regard the pooled representation as the ``image embedding vector'' $\bm{x}_1$. Then, $\bm{x}_1$ will be fed as the input at time step~1 into the RNN-based hierarchical module. Whilst, at the fine-grained classification level (\emph{i.e.}, vehicle-level), these deep descriptors plus the intermediate output of the RNN-based hierarchical module will be used for generating the attended descriptors with the following attention embedding vector $\bm{x}_2$, cf. Fig.~\ref{fig:pipeline}.

\subsection{Crucial modules in RNN-HA}

In the following sections, we elaborate the RNN-based hierarchical module and the attention module which are two crucial modules of our RNN-HA model.

\subsubsection{RNN-based hierarchical module}

Recurrent Neural Network (RNN)~\cite{rnn} is a class of neural network that maintains internal hidden states to model the dynamic temporal behavior of sequences with arbitrary lengths through directed cyclic connections between its units. It can be considered as a hidden Markov model extension that employs non-linear transition function and is capable of modeling long/short term temporal dependencies.

We hereby employ RNN to capture the latent hierarchical label dependencies existing in vehicle re-identification. Concretely, we choose Gated Recurrent Units (GRUs)~\cite{gru14EMNLP} as the gating mechanism to implement RNN. GRUs have fewer parameters than another popular gating mechanism, \emph{i.e.}, LSTM~\cite{rnn}, which makes GRU much easier to optimize. Moreover, evaluations by Chung \emph{et al.}~\cite{chungarXiv} found that when LSTM and GRU have the same amount of parameters, GRU slightly outperforms LSTM. Similar observations were also corroborated in~\cite{rnn15ICML}.

GRU contains two gates: an update gate $\bm{z}$ and a reset gate $\bm{r}$. We follow the model used in~\cite{gru14EMNLP}. Let $\sigma$ be the sigmoid non-linear activation function. The GRU updates for time step $t$ given inputs $\bm{x}_t$, $\bm{h}_{t-1}$ are:
\begin{align}
\bm{z}_t &= \sigma(W_{xz}\bm{x}_t+W_{hz}\bm{h}_{t-1}+b_z) \,,\\
\bm{r}_t &= \sigma(W_{xr}\bm{x}_t+W_{hr}\bm{h}_{t-1}+b_r) \,,\\
\bm{n}_t &= {\tanh}(W_{xg}\bm{x}_t+\bm{r}_t \odot W_{hg}\bm{h}_{t-1}+b_g) \,,\\
\bm{h}_t &= (1-\bm{z}_t)\odot \bm{n}_t + \bm{z}_t \odot \bm{h}_{t-1} \,.
\end{align}
Here, $\odot$ represents the product with a gate value, and various $W$ matrices are learned parameters.

As shown in Fig.~\ref{fig:pipeline}, we decompose the coarse-to-fine hierarchical classification problem into an ordered prediction path, \emph{i.e.}, from car model to specific vehicle. The prediction path can reveal the hierarchical characteristic beneath this two-stage classification problem. Meanwhile, the probability of a path can be computed by the RNN model. 

As aforementioned, since we aim to solve coarse-grained classification at the first step, global image features which represent global visual information are required. Thus, the image embedding vector $\bm{x}_1$ produced by simply global average pooling is used as the input at the first time step ($t=1$) in the hierarchical module. Then, the output vector $\bm{o}_1$ at $t=1$ is employed for computing the coarse-grained (model-level) classification loss, \emph{i.e.}, $\mathcal{L}_{\rm model}$. At the same time, $\bm{o}_1$ will be transformed via a transformer network (cf. the green sub-module in Fig.~\ref{fig:pipeline}) into an attention guidance signal $\bm{w}$. The attention signal can guide the subsequent attention network to learn which deep descriptors should be attended when identifying different specific vehicles. The details of the attention network are presented in the next sub-section.

Now suppose we obtain a well-trained attention network, it could focus on these descriptors corresponding to subtle discriminative image regions (\emph{e.g.}, customized paintings, favorite decorations, etc.), and neglect these descriptors corresponding to common patterns (\emph{e.g.}, the similar headlights, car roofs, etc.). Based on the attentions on descriptors, we can obtain the attention embedding vector $\bm{x}_2$ which is also the input of the RNN-based hierarchical module at $t=2$. The rest procedure at time step~2 is computing the loss in fine-grained classification (\emph{i.e.}, vehicle-level classification) based on the output vector $\bm{o}_2$.

At last, the final loss of our RNN-HA is formed by the summation of both the coarse-grained (\emph{i.e.}, model-level) and fine-grained (\emph{i.e.}, vehicle-level) classification loss as:
\begin{equation}
\mathcal{L} = \mathcal{L}_{\rm model} + \mathcal{L}_{\rm vehicle}\,.
\end{equation}
In our implementation, the traditional cross entropy loss function is employed in each loss branch. Note that, there is no tuning parameter in this equation, which reveals RNN-HA is generalized and not tricky.

\subsubsection{Attention module}

After performing the coarse-grained classification, it is required to identify different specific vehicles at this fine-grained level. What distinguishes different vehicles from each other is the subtle visual information, such as customized paintings, windshield stickers, favorite decorations, etc. To capture these subtle but important cues, we propose to use an attention model to focus processing only on the attended deep descriptors reflecting those cues. Meanwhile, discarding those descriptors corresponding to common patterns, \emph{e.g.}, the same car roofs of the same one car model, is also crucial for identifying specific vehicles.

To achieve these goals, we rely on the output $\bm{o}_1$ of the RNN-based hierarchical module at $t=1$ by regarding it as the guidance signal for the subsequent attention learning. Specifically, we design a transformer network to transform $\bm{o}_1$ into a new space where it could play a role as the attention guidance signal $\bm{w}$. The transformer network contains two fully connected layers with a ReLU layer between them. After this transforming, $\bm{w}$ is utilized for evaluating which deep descriptors should be attended or overlooked.

Given an input image, ${\bm{f}}_{(i,j)} \in \mathbb{R}^d$ is the deep descriptor in the $(i,j)$ spatial location in the activation tensor of the last convolutional layer, where $i\in \left\{1,2,\ldots,h\right\}$ and $j\in \left\{1,2,\ldots,w\right\}$. Based on the attention guidance signal $\bm{w}$, we can get the corresponding unnormalized attention scores $s_{(i,j)} \in \mathbb{R}^1$ of ${\bm{f}}_{(i,j)}$ by:
\begin{equation}
\label{eq:atten}
s_{(i,j)}=g(\bm{w}^\top {\bm{f}}_{(i,j)}) \,,
\end{equation}
where $g(\cdot)$ is the softplus function, \emph{i.e.}, $g(x)=\ln (1+\exp(x))$. Since we only aim to concern with the relative importance of the deep descriptors within an image, we further normalize the attention scores into the $\left[0,1\right]$ interval for aggregating the descriptors:
\begin{equation}
\label{eq:att}
a_{(i,j)} = \frac{s_{(i,j)}+\epsilon}{\sum_i\sum_j(s_{(i,j)}+\epsilon)} \,,
\end{equation}
where $a_{(i,j)}$ is the normalized attention score, and $\epsilon$ is a small constant set to $0.1$ in our experiments.

Thus, different deep descriptors will receive the corresponding (normalized) attention scores depending on how much attention paid to itself. Finally, we can get the attended feature representation $\hat{\bm{f}}_{(i,j)} \in \mathbb{R}^d$ by applying $a_{(i,j)}$ to ${\bm{f}}_{(i,j)}$ as follows:
\begin{equation}
\hat{\bm{f}}_{(i,j)} = a_{(i,j)}\odot {\bm{f}}_{(i,j)}\,,
\end{equation}
where $\odot$ is the element-wise multiplication. Then, after global average pooling employed in $\hat{\bm{f}}_{(i,j)}$, we can obtain the attention embedding vector:
\begin{equation}
\bm{x}_2 = \frac{1}{h\times w} \sum_{i}\sum_{j} \hat{\bm{f}}_{(i,j)} \,.
\end{equation}
Then, $\bm{x}_2 \in \mathbb{R}^d$ will be fed as the input into the RNN-based hierarchical module at $t=2$. At last, by treating different vehicles as classification categories, the output $\bm{o}_2$ at $t=2$ is used for recognizing vehicles at the fine-grained level. As each module is differentiable, the whole RNN-HA model is end-to-end trainable.

Additionally, in theory, the proposed RNN-based hierarchical module can be composed of more time steps, \emph{i.e.}, $t>2$. Since the vehicle datasets used in experiments have two-level hierarchical labels, we use our RNN-HA with $t=2$ for the vehicle re-identification problem. Beyond that, the proposed RNN-HA model can also deal with the hierarchical fine-grained recognition problem, \emph{e.g.},~\cite{fg17ICCV}, which reveals the generalization usage of our model.

\section{Experiments}\label{sec:experiment}

In this section, we first describe the datasets and evaluation metric used in experiments, and present the implementation details. Then, we report vehicle re-identification results on two challenging benchmark datasets.

\subsection{Datasets and evaluation metric}

To evaluate the effectiveness of our proposed RNN-HA model, we conduct experiments on two benchmark vehicle re-identification datasets, \emph{i.e.}, \emph{VeRi}~\cite{fact16ICME} and \emph{VehicleID}~\cite{vehicleID16CVPR}.

The \emph{VeRi} dataset contains more than $50,000$ images of $776$ vehicles with identity annotations and car types. The car types in \emph{VeRi} include the following ten kinds: ``sedan'', ``SUV'', ``van'', ``hatchback'', ``MPV'', ``pickup'', ``bus'', ``truck'' and ``estate''. The \emph{VeRi} dataset is split into a training set consisting of $37,778$ images of $576$ vehicles and a test set of $11,579$ images belonging to $200$ vehicles. In the evaluation, a subset of $1,678$ images in the test set are used as the query images. The test protocol in \cite{fact16ICME,provid16ECCV} recommends that evaluation should be conducted in an image-to-track fashion, in which the image is used as the query, while the gallery consists of tracks of the same vehicle captured by other cameras. The mean average precision (mAP), top-1 and top-5 accuracy (CMC) are chosen as the evaluation metric for the \emph{VeRi} dataset.

Compared with \emph{VeRi}, \emph{VehicleID} is a large-scale vehicle re-identification dataset, which has a total of $26,267$ vehicles with $221,763$ images, and $10,319$ vehicles are labeled with models such as ``Audi A6'', ``Ford Focus'', ``Benz GLK'' and so on. There are $228$ car models in total. \emph{VehicleID} is split into a training set with $110,178$ images of $13,134$ vehicles and a testing set with $111,585$ images of $13,133$ vehicles. For \emph{VehicleID}, the image-to-image search is conducted because each vehicle is captured in one image by one camera. For each test dataset (size = $800$, $1,600$ and $2,400$), one image of each vehicle is randomly selected into the gallery set, and all the other images are query images. We repeat the above processing for ten times, and report the average results. Following the previous work~\cite{vehicleID16CVPR}, the evaluation metric for \emph{VehicleID} is top-1, top-5 accuracy (CMC).

\subsection{Implementation details}

In our experiments, vehicles' identity annotations of both datasets are treated as the vehicle-level classification categories, while the car types of \emph{VeRi} and the car models of \emph{VehicleID} are regarded as the model-level classification categories for these two datasets, respectively. Note that, for both datasets, the vehicle-level classification categories of training and testing are disjoint. All images are of $224\times 224$ image resolution. After training RNN-HA, we employ our model as a feature extractor for extracting test image features. Specifically, the output $\bm{o}_2$ at time step 2 of our RNN-HA is the acquired image representation. In the evaluation, $\bm{o}_2$ of test images are firstly $\ell_2$-normalized, and then the cosine distance acts as the similarity function in re-identification.

For fair comparisons, following~\cite{vehicleID16CVPR}, we adopt \textit{VGG\_CNN\_M\_1024}~\cite{VGGarXiv} as the base model of our representation learning module. For the RNN-based module, the number of hidden units in GRU is $1,024$, and the zero vector is the hidden state input at time step 0. During training, we train our unified RNN-HA in an end-to-end manner by employing the RMSprop~\cite{rmsprop} optimization method with its default parameter settings to update model parameters. The learning rate is set to $0.001$ at the beginning, and five epochs later, it is reset to $0.0001$. The batch size is $64$. We implement our model by the open-source library PyTorch.


\subsection{Main results}

We present the main vehicle re-identification results by firstly introducing state-of-the-arts and our baseline methods, then following the comparison results on two datasets.\footnote{Note that, in Table~\ref{table:veriresult} and Table~\ref{table:pkuresult}, we only compare with the appearance-based methods for fair comparisons.}

\subsubsection{Comparison methods}

We compare nine vehicle re-identification approaches which are evaluated on \emph{VehicleID} and \emph{VeRi}. The details of these approaches are introduced as follows. Among them, many state-of-the-art methods, \emph{e.g.},~\cite{vehicleID16CVPR,bmvc17Vid,CVPR18carreid}, also employ both vehicle-level and model-level supervision, which has the same formation as ours.
\begin{itemize}
\item \textit{Local Maximal Occurrence Representation (LOMO)}~\cite{LOMO} is the state-of-the-art text features for person re-identification. We follow the optimal parameter settings given in \cite{LOMO} when adopting LOMO on these two vehicle re-identification datasets.
\item \textit{Color based feature (BOW-CN)}~\cite{bowcn15ICCV} is a benchmark method in person re-identification, which applies the Bag-of-Words (BOW) model~\cite{jegouvlad2010,jorgeiccv13} with Color Name (CN) features~\cite{color09TIP}. By following \cite{bowcn15ICCV}, a $5,600$-d BOW-CN feature is obtained as the color based feature.
\item \textit{Semantic feature learned by CNN (GoogLeNet)} adopts the GoogLeNet model~\cite{googlenet} which is fine-tuned on the CompCars dataset~\cite{compcar15CVPR} as a powerful feature extractor for high-level semantic attributes of the vehicle appearance. The image feature of this method is $1,024$-d extracted from the last pooling layer of GoogLeNet.
\item \textit{Fusion of Attributes and Color feaTures (FACT)}~\cite{fact16ICME} is proposed for vehicle re-identification by combining deeply learned visual feature from GoogLeNet~\cite{googlenet}, BOW-CN and BOW-SIFT feature to measure only the visual similarity between pairs of query images.
\item \textit{Siamese-Visual}~\cite{cuhkcar17ICCV} relys on a siamese-based neural network which has a symmetric structure to learn the similarity between a query image pair with appearance information.
\item \textit{Triplet Loss}~\cite{triplossPR} is adopted in vehicle re-identification by learning a harmonic embedding of each input image in the Euclidean space that tends to maximize the relative distance between the matched pair and the mismatched pair.
\item \textit{Coupled Cluster Loss (CCL)}~\cite{vehicleID16CVPR} is proposed for improving triplet loss by replacing the single positive/negative input sample by positive/negative images sets, which can make the training phase more stable and accelerate the convergence speed.
\item \textit{Mixed Diff+CCL}~\cite{vehicleID16CVPR} adopts a mixed network structure with a coupled cluster loss to learn the relative distances of different vehicles. Note that, the most different point between it and our model is that it used these two level categories in a simple parallel fashion, while we formulate them into a hierarchical fashion. Experimental results could validate the effectiveness of our proposed hierarchical model, and justify the hierarchical fashion should be the optimal option.
\item \textit{CLVR}~\cite{bmvc17Vid} consists of two branches in the cross-modality paradigm, where one branch is for the vehicle model level and the other is for the vehicle ID level. Again, the two level categories in their model are not formulated into a hierarchical fashion like ours.
\item \textit{VAMI}~\cite{CVPR18carreid} proposes a viewpoint-aware attentive multi-view inference model by leveraging the cues of multiple views to deal with vehicle re-identification.
\end{itemize}

Additionally, to investigate the impacts of the various modules in our end-to-end framework, we analyze the effects of the RNN-based hierarchical module and the attention module by conducting experiments on two baselines:
\begin{itemize}
\item \textit{FC-HA (w/o RNN)} replaces the RNN hierarchical module by simply employing traditional fully-connected layers as direct transformation, but keeps the attention mechanism.
\item \textit{RNN-H w/o attention} keeps the hierarchical module, but removes the attention module from our proposed RNN-HA model. Specifically, the inputs $\bm{x}_1$ and $\bm{x}_2$ at $t=1$ and $t=2$ are both the representations global average pooled by these deep convolutional descriptors.
\end{itemize}

\subsubsection{Comparison results on \emph{VeRi}}

Table~\ref{table:veriresult} presents the comparison results on the \emph{VeRi} dataset. Our proposed RNN-HA model achieves $52.88\%$ mAP, $77.03\%$ top-1 accuracy and $90.91\%$ top-5 accuracy on \emph{VeRi}, which significantly outperforms the other state-of-the-art methods. These results validate the effectiveness of the proposed model. Moreover, RNN-HA has a gain of $5.69\%$ mAP and $15.47\%$ top-1 accuracy comparing with the FC-HA baseline method, which proves the effectiveness of the RNN-based hierarchical design. Also, compared with ``RNN-H w/o attention'', RNN-HA achieves a gain of $3.96\%$ mAP and $11.75\%$ top-1 accuracy. It justifies our proposed attention module when identifying different specific vehicles at the fine-grained classification level.

In addition, to further improve the re-identification accuracy, we simply replace the \textit{VGG\_CNN\_M\_1024} model of the representation learning module with \emph{ResNet-50}~\cite{kaiming15residual}. Our modified model is denoted as ``RNN-HA (ResNet)'', which obtains $56.80\%$ mAP, $80.79\%$ top-1 accuracy and $92.31\%$ top-5 accuracy on \emph{VeRi}.

\begin{table}[t]
 \caption{Comparison of different methods on \emph{VeRi}~\cite{fact16ICME}.} \label{table:veriresult}
 \centering
 \small
 \begin{tabular}{|c|c|c|c|}
  \hline
	Methods & mAP & Top-1 & Top-5 \\
  \hline
  \hline
        LOMO~\cite{LOMO} & 9.64 & 25.33 & 46.48 \\
	BOW-CN~\cite{bowcn15ICCV} & 12.20 & 33.91  & 53.69  \\
	GoogLeNet~\cite{compcar15CVPR} & 17.89 & 52.32 & 72.17  \\
	FACT~\cite{fact16ICME} & 18.75 & 52.21  & 72.88  \\
	Siamese-Visual~\cite{cuhkcar17ICCV} & 29.48 & 41.12  & 60.31  \\
	VAMI~\cite{CVPR18carreid} & 50.13 & 77.03 & 90.82 \\
  \hline
  \hline
        FC-HA (w/o RNN) & 47.19 & 61.56 & 76.88 \\
        RNN-H w/o attention & 48.92 & 63.28  & 78.82  \\
  \hline
	Our RNN-HA  & 52.88 & 66.03  & 80.51  \\
	Our RNN-HA (ResNet)  & \textbf{56.80} &   \textbf{74.79} &  \textbf{87.31} \\
  \hline
 \end{tabular}
\end{table}

\subsubsection{Comparison results on \emph{VehicleID}}

\begin{table*}[t]
 \caption{Comparison of different methods on the \emph{VehicleID} dataset~\cite{vehicleID16CVPR}.} \label{table:pkuresult}
 \centering
 \small
\begin{tabular}{|c|c|c|c|c|c|c|}
  \hline
  \multirow{2}{*}{Methods} & \multicolumn{2}{c|}{Test size = $800$} & \multicolumn{2}{c|}{Test size = $1,600$} & \multicolumn{2}{c|}{Test size = $2,400$} \\
  \cline{2-7} & Top-1 & Top-5 & Top-1 & Top-5 & Top-1 & Top-5 \\
  \hline
  \hline
    LOMO~\cite{LOMO} & 19.7 & 32.1 & 18.9 & 29.5 & 15.3 & 25.6 \\
    BOW-CN~\cite{bowcn15ICCV} & 13.1 & 22.7 & 12.9 & 21.1 & 10.2 & 17.9 \\
    GoogLeNet~\cite{compcar15CVPR} & 47.9 & 67.4 & 43.5 & 63.5 & 38.2 & 59.5 \\
    FACT~\cite{fact16ICME} & 49.5 & 67.9 & 44.6 & 64.2 & 39.9 & 60.5 \\
    Triplet Loss~\cite{triplossPR} & 40.4 & 61.7 & 35.4 & 54.6 & 31.9 & 50.3 \\
    CCL~\cite{vehicleID16CVPR} & 43.6 & 64.2 & 37.0 & 57.1 & 32.9 & 53.3 \\
    Mixed Diff+CCL~\cite{vehicleID16CVPR} & 49.0 & 73.5 & 42.8 & 66.8 & 38.2 & 61.6 \\
    {CLVR}~\cite{bmvc17Vid} & 62.0 & 76.0 & 56.1 & 71.8 & 50.6 & 68.0 \\
    VAMI~\cite{CVPR18carreid} & 63.1 & 83.3 & 52.8 & 75.1 & 47.3 & 70.3 \\
  \hline
  \hline
    FC-HA (w/o RNN) & 56.7 & 74.5 & 53.6 & 70.6 & 48.6 & 66.3 \\ 
    RNN-H w/o attention & 64.5 & 78.8 & 62.4 & 75.9 & 59.0 & 74.2 \\ 
  \hline
    Our RNN-HA & 68.8 & 81.9 & 66.2 & 79.6 & 62.6 & 77.0 \\ 
    Our RNN-HA (672) & 74.9 & 85.3 & 71.1 & 82.3 & 68.0 & 81.4 \\ 
    Our RNN-HA (ResNet+672) & \textbf{83.8} & \textbf{88.1} & \textbf{81.9} & \textbf{87.0} & \textbf{81.1} & \textbf{87.4} \\ 
  \hline
 \end{tabular}
\end{table*}

For the large-scale dataset, \emph{VehicleID}, we report the comparison results in Table~\ref{table:pkuresult}. On different test settings (\emph{i.e.}, test size = $800$, $1,600$ and $2,400$), our proposed RNN-HA achieves the best re-identification performance on this large-scale dataset. An interesting observation in both tables is that the FC-HA baseline method outperforms all the Siamese or triplet training methods on both \emph{VeRi} and \emph{VehicleID}. It is consistent with the observations in recently most successful person re-identification approaches, \emph{e.g.},~\cite{surveyarXiv,zhedongarXiv}. These approaches argue that a classification loss is superior for the re-identification task, while the triplet loss or siamese-based nets perform unsatisfactorily due to its tricky training example sampling strategy.

\begin{figure}[t!]
\centering
	{\includegraphics[width=0.8\columnwidth]{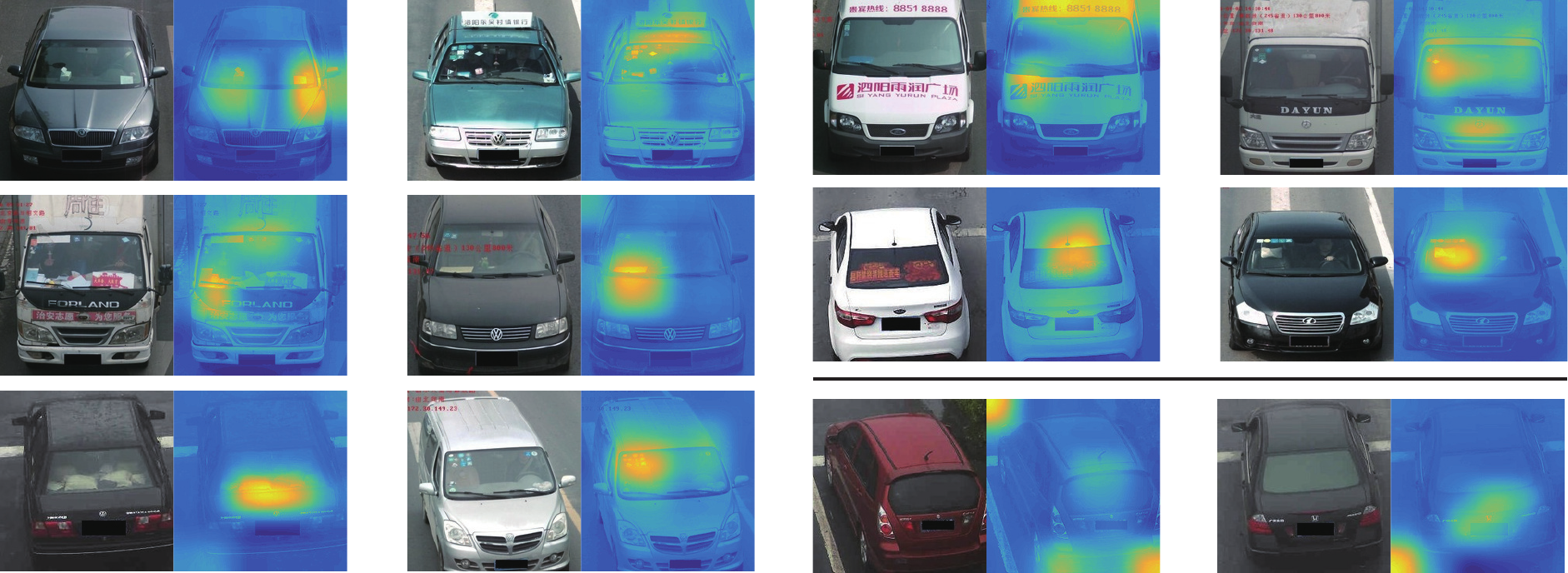}}
\caption{Examples of the attention maps on \emph{VehicleID}. The brighter the region, the higher the attention scores. (Best viewed in color and zoomed in.)}
\label{fig:att}
\end{figure}

\begin{figure*}[t!]
\centering
	{\includegraphics[width=\textwidth]{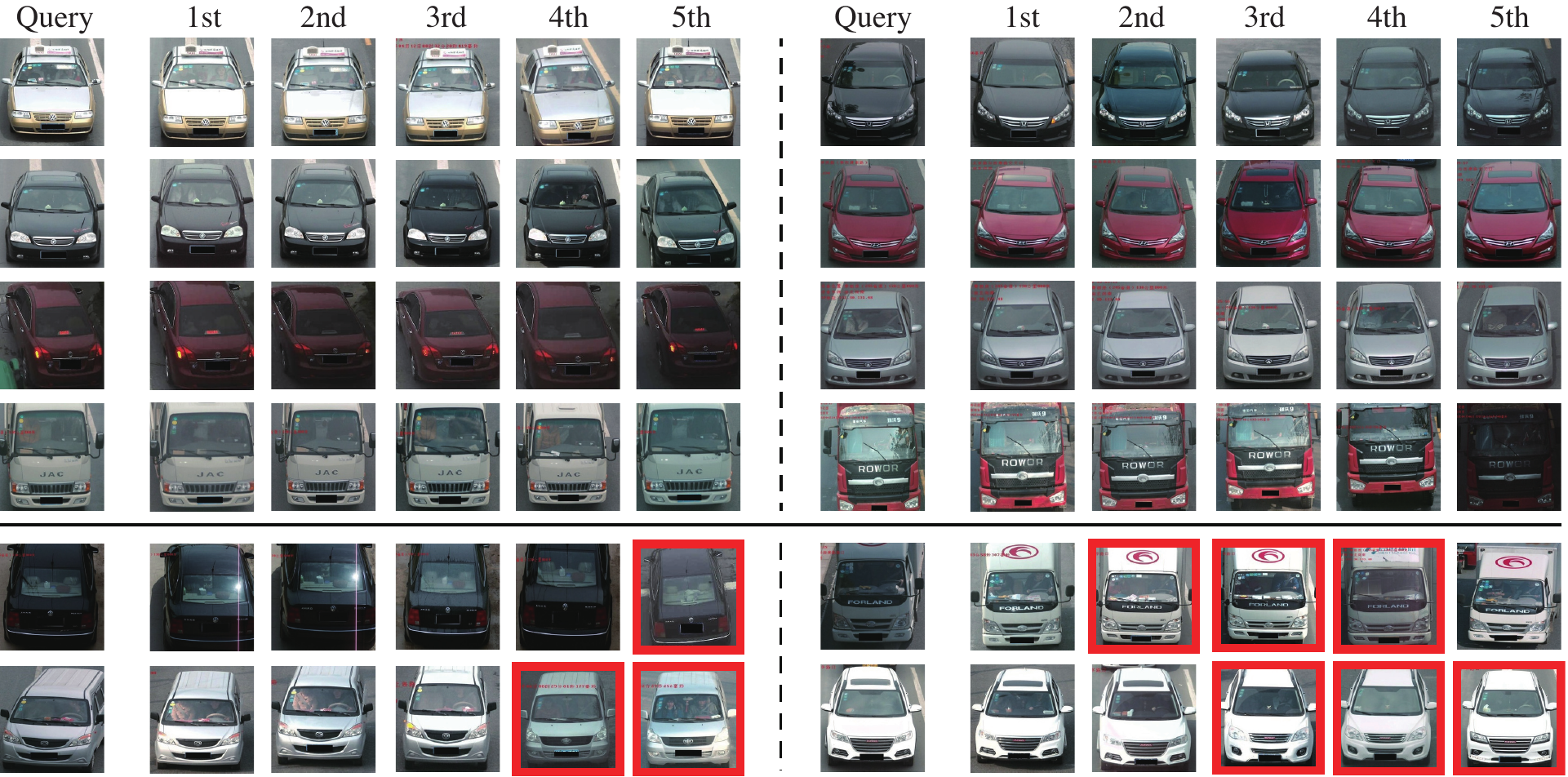}}
\caption{The top-5 re-identification results on the large-scale \emph{VehicleID} dataset. The upper eight examples are successful, while the lower four examples are failure cases. Red boxes denote false positives. (Best viewed in color and zoomed in.)}
\label{fig:reid}
\end{figure*}

From the qualitative perspective, Fig.~\ref{fig:att} shows the learned attention maps (\emph{i.e.}, $a_{(i,j)}$ in Eq.~\ref{eq:att}) of several random sampled test vehicle images. We can find that the attended regions accurately correspond to these subtle and discriminative image regions, such as windshield stickers, stuffs placed behind windshield or rear windshield, and customized paintings. In addition, Fig.~\ref{fig:reid} presents several re-identification results returned by our RNN-HA on \emph{VehicleID}.

Furthermore, on this large-scale dataset, we also use input images with a high image resolution, \emph{i.e.}, $672\times 672$, since higher resolution could benefit to learn a more accurate attention map. The RNN-HA model with the $672\times 672$ resolution is denoted by ``RNN-HA (672)''. Apparently, it improves the re-identification top-1 accuracy by $5\sim 6\%$. Besides, based on the high resolution, we also modify RNN-HA by equipping it with \emph{ResNet-50}, and report its results as ``RNN-HA (ResNet+672)'' in Table~\ref{table:pkuresult}. On such challenging large-scale dataset, even though we only depend on appearance information, our model could achieve $83.8\%$ top-1 accuracy, which reveals its effectiveness in real-life applications.

\section{Conclusion}\label{sec:conclude}

In this paper, we noticed there is a coarse-to-fine hierarchical dependency natively beneath the vehicle re-identification problem, \emph{i.e.}, from coarse-grained (car models/types) to fine-grained (specific vehicles). To model this important hierarchical dependent relationship, we proposed a unified RNN-based hierarchical attention (RNN-HA) model consisting of three mutually coupled modules. In RNN-HA, after obtaining the deep convolutional descriptors generated by the first representation learning module, we leveraged the powerful RNNs and further designed a RNN-based hierarchical module to mimic such hierarchical classification process. Moreover, at the fine-grained level, the attention module was developed to capture subtle appearance cues for effectively distinguish different specific vehicles.

On two benchmark datasets, \emph{VeRi} and \emph{VehicleID}, the proposed RNN-HA model achieved the best re-identification performance. Besides, extensive ablation studies demonstrated the effectiveness of individual modules of RNN-HA. In the future, it is promising to further improve the re-identification performance by incorporating more attributes information into our proposed RNN-HA framework.

\bibliographystyle{splncs04}
\bibliography{reid}

\end{document}